\pdfoutput=1

\documentclass[11pt]{article}

\usepackage[final]{acl}

\usepackage{times}
\usepackage{latexsym}

\usepackage[T1]{fontenc}

\usepackage[utf8]{inputenc}

\usepackage{microtype}

\usepackage{inconsolata}

\usepackage{graphicx}

\title{Efficacy of ByT5 in Multilingual Translation of\\Biblical Texts for Underrepresented Languages}

\author{Corinne Aars$^\dag$ \quad Lauren Adams$^\dag$ \quad Xiaokan Tian$^\dag$ \quad Zhaoyu Wang$^\dag$ \quad Colton Wismer$^\dag$ \\ \bf Jason Wu$^\dag$ \quad Pablo Rivas$^*$ \quad Korn Sooksatra \quad Matthew Fendt \\
  Department of Computer Science \quad Baylor University \\ $^\dag$Equal Contribution \quad 
  $^*$Correspondence: \texttt{Pablo\_Rivas@Baylor.edu}}

\begin{document}
\maketitle
\begin{abstract}
This study presents the development and evaluation of a ByT5-based multilingual translation model tailored for translating the Bible into underrepresented languages. Utilizing the comprehensive Johns Hopkins University Bible Corpus, we trained the model to capture the intricate nuances of character-based and morphologically rich languages. Our results, measured by the BLEU score and supplemented with sample translations, suggest the model can improve accessibility to sacred texts. It effectively handles the distinctive biblical lexicon and structure, thus bridging the linguistic divide. The study also discusses the model’s limitations and suggests pathways for future enhancements, focusing on expanding access to sacred literature across linguistic boundaries.
\end{abstract}

\section{Introduction}
This study aims to use an advanced language model to make the translation of sacred texts, like the Bible, into less commonly spoken languages more efficient and quicker. The Bible's structured format, with its division into books, chapters, and verses, allows for a wide variety of uses and highlights its rich foundational vocabulary. Using this extensive corpus, we trained a multilingual translation model to produce translations of Bible verses for languages that lack representation~\citep{peters-etal-2018-deep}.

At present, translating religious texts into less commonly spoken languages is a lengthy and complex process. Finding a translator with proficiency in both the source and target languages can be challenging. Often, this translator is the primary person working on the translation, which can take a considerable amount of time to complete even a first draft. The subsequent process of refining this draft can extend over many years, delaying the availability of these important texts in the native languages of various communities. Our model aims to complement rather than replace the traditional human translation process by making it more streamlined.

This research leverages the ByT5 multilingual translation model, trained on the John Hopkins University Bible Corpus \citep{xue2022byt5,mccarthy-etal-2020-johns}, to facilitate the translation of Bible verses into several underrepresented languages. The ByT5 model is celebrated for its ability to produce high-quality text in numerous languages \citep{JMLR:v21:20-074}, offering a robust solution to the challenges of translating sacred texts into languages with few resources. Our innovative method seeks to mitigate the issues inherent in conventional translation practices, such as the lack of available skilled translators and long project timelines. By employing the ByT5 model, we aim to enhance the speed and precision of Bible translations for languages with scant resources and expertise. Integrating advanced technology with linguistic diversity, we aspire to broaden the accessibility and cultural preservation of sacred texts, thereby enriching the cultural and linguistic fabric of religious literature for communities speaking underrepresented languages.

\section{Background}

ByT5 is an extension of the T5 model, which itself is a language model recognized for its effectiveness and was developed by Google Research~\citep{JMLR:v21:20-074}. ByT5 improves upon T5 by adopting byte-level tokenization, which enhances its proficiency with character-based languages such as Chinese and Japanese.

The tokenization at the byte level allows ByT5 to handle scripts that are character-intensive and may not exhibit clear word boundaries, or which utilize characters more extensively than words. This feature makes ByT5 particularly adept at interpreting the intricacies of such languages~\citep{xue2022byt5}.

For languages characterized by rich morphology, where words can have various inflections and derivations, ByT5's tokenizer has shown increased accuracy, capturing morphological variations more precisely~\citep{fujii2023different}. These capabilities are beneficial for translation, summarization, and question-answering tasks. By tokenizing at the character level, ByT5 demonstrates enhanced generalizability across languages, which is particularly useful for languages with sparse training data. This adaptability has been documented to improve multilingual model performance. Furthermore, byte-level tokenization generally results in a smaller vocabulary than word-level tokenization, which contributes to more efficient model training and inference, as well as reduced memory and computational needs.


The John Hopkins University Bible Corpus was selected to train our model. This corpus is notable for its size and organization, comprising over 4000 versions of the Bible across more than 1600 languages~\citep{mccarthy-etal-2020-johns}. Its parallel structure across translations makes it advantageous for machine learning applications. Other corpora fall short in these particular aspects~\citep{Sierra.2024}.

This corpus is especially marked by its linguistic diversity, offering both full and partial texts in a myriad of languages, some of which are significantly underrepresented with only a single book of the Bible translated. Such disparity highlights the necessity of our work.

The construction of this corpus involved extensive web scraping and merging with existing corpora. It underwent a thorough cleaning and alignment process to ensure the texts were structured verse-parallel, which is ideal for training machine learning models.

\section{Methodology} 

This section delineates the systematic approach employed for training our model and tuning its parameters.

\subsection{Model Training}

The Byt5 model was selected for its proficiency in byte-level tokenization, which is critical for discerning subtle linguistic differences in underrepresented languages~\citep{wang2020neural}. Our primary dataset, derived from the Johns Hopkins Bible Corpus, encompasses various linguistic variations, furnishing our models with an extensive linguistic foundation for learning~\citep{mccarthy-etal-2020-johns}.

\subsection{Parameter Tuning}

We determined the ByT5 training hyper-parameters through an iterative experimental protocol~\citep{ebrahimi-kann-2021-adapt}. Adjustments were made to the learning rate, early stop criteria, patience, and batch size, to distill an optimal parameter set for our specific translation endeavor. Our configuration included a learning rate of 0.0002, a scheduler factor of 0.5, patience of 10, a batch size of 48, and a maximum of 500 epochs with early stopping that usually ceases around epoch 43. Our dataset encompassed 3 million data samples, i.e., pairs of source-target translations.

By coupling a linguistically diverse corpus with meticulous parameter optimization, we aim to bolster ByT5's translation efficacy for languages that are often overlooked in machine translation~\citep{costa2022no}.

\section{Results and Discussion}

Moving from our methodology to its outcomes, we evaluate the translation quality of our ByT5-based multilingual translation model.

\subsection{Translation Quality}

We utilized the BLEU (Bilingual Evaluation Understudy) score, a standard metric in machine translation evaluation, to assess our model's accuracy by comparing its outputs with human translations. Our model achieved a BLEU score of 0.27, which signals its potential in handling translations, particularly in the domain of underrepresented languages.

The score achieved not only sheds light on the model's capabilities but also underscores the intricate nature of machine translation tasks. Although the BLEU score is a valuable metric, it does not capture all aspects of language translation, such as the true linguistic fidelity and fluency that are crucial in the context of less common languages, which often lack established benchmarks.

For a more comprehensive illustration of our model's performance across various language pairs, we have compiled a series of translation examples in Table~\ref{tbl:samples}.

\begin{table*}[h!]
\centering
\includegraphics[width=\textwidth, trim=1cm 12.9cm 1.3cm 3.5cm, clip]{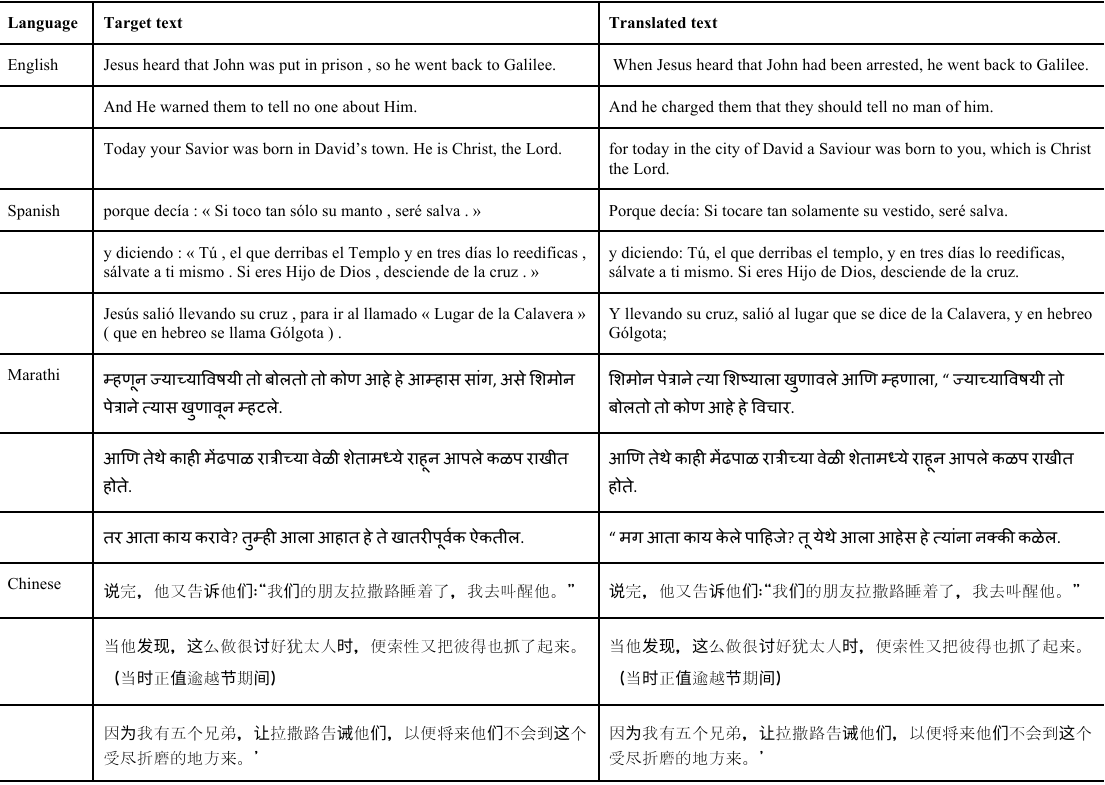}
\caption{Comparative Analysis of ByT5 Model Translations with Target Texts in Multiple Languages. In this example, Marathi is an underrepresented language. Please see Table~\ref{tbl:samples_ap} in the Appendix for more samples.}
\label{tbl:samples}
\end{table*}

\subsection{Analysis}

The outcomes presented above offer a glimpse into a promising avenue for employing NLP models in assisting with the translation of sacred texts. Our model's particular strength lies in its fine-tuning to the specificities of biblical language and style. Contrary to broader language models like Google Translate, our specialized training on the biblical corpus allows our system to adeptly handle the unique linguistic features of widely spoken and underrepresented languages alike.

The application of NLP to support the translation of significant texts holds promise for broader cultural and linguistic access. Such improvements in translation technology could hasten the delivery of these important works to communities with underrepresented languages. Our findings demonstrate the potential and scalability of language models to translate a vast spectrum of sacred texts. This model shows the capacity of NLP to bridge linguistic and cultural divides. As NLP technologies advance, they herald new possibilities for promoting linguistic diversity and spreading knowledge.

\section{Conclusions}
The investigation into the ByT5 model's ability to translate sacred texts into underrepresented languages has demonstrated promising results. Our research revealed that the model, trained on the Johns Hopkins University Bible Corpus, not only can handle the complexities inherent to diverse linguistic structures but also shows potential in improving access to significant cultural literature. Despite the challenges underscored by the BLEU scores, which may not fully capture linguistic nuances, the model's translations remained reasonably accurate and faithful to the source texts.

Future research could explore refining the model's parameters further to enhance translation quality, especially for languages with limited linguistic data. Additionally, the integration of ByT5 with other NLP tools may yield better comprehension of context and idiomatic expressions. The overarching aim of our work—to facilitate the sharing of sacred texts across cultural and linguistic barriers—has been substantiated by the findings, suggesting that advanced NLP models can indeed be a catalyst for increased inclusivity in the realm of sacred literature. The scalability of our approach also opens avenues for extending this work to other significant texts, potentially enriching the cultural and educational resources available to underrepresented language speakers worldwide.

\section*{Acknowledgments}

Part of this work was funded by the National Science Foundation under grants CNS-2210091, CHE-1905043, and CNS-2136961.

\bibliography{custom}

\appendix

\section{Appendix: Additional Sample Translations}
\label{sec:appendix}

This appendix features Table~\ref{tbl:samples_ap}, which showcases the ByT5 model's translation capabilities across several languages. Each entry includes the original target text followed by the translated text generated by the model. The juxtaposition illustrates the model's proficiency in maintaining the semantic integrity of complex religious narratives, as well as its handling of diverse linguistic structures ranging from English and German to more character-centric languages like Chinese and Japanese. This comparative exhibit serves not only as a testament to the ByT5 model's linguistic versatility but also as a valuable resource for future research and development in the field of machine translation of sacred texts.

\begin{table*}[t!]
\centering
\includegraphics[width=\textwidth, trim=1cm 12.5cm 1.3cm 2.5cm, clip]{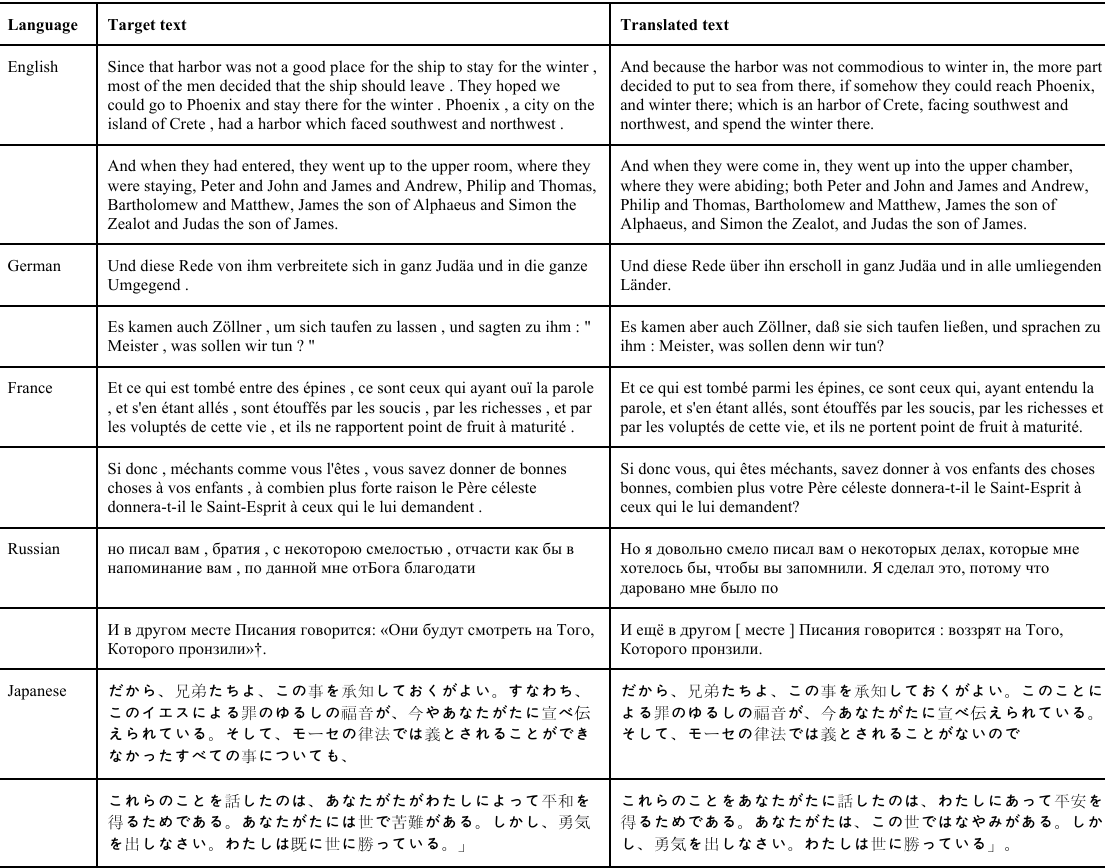}
\caption{Comparative Analysis of ByT5 Model Translations with Target Texts in Multiple Languages.}
\label{tbl:samples_ap}
\end{table*}

\end{document}